\documentclass[]{article}
\usepackage{proceed2e}

\usepackage{times}
\usepackage{graphicx}
\usepackage{amsmath}
\usepackage{amssymb}
\usepackage{paralist}
\usepackage{color}
\usepackage{url}
\usepackage[round,sort&compress]{natbib}
\usepackage{pdflscape}
\usepackage{multicol}
 
\newcommand{\defoccur}[1]{\textsl{#1}}
\newcommand{\green}{\color{green}}

\usepackage{xspace}
\def\onedot{\ifx\@let@token.\else.\null\fi\xspace}
\def\eg{{e.g}\onedot} 
 
\def\cf{{c.f}\onedot} 
\def\etc{{etc}\onedot}
\def\vs{{vs}\onedot}

\def\etal{{et al}\onedot}

\begin{document}

\title{Large-Scale Automatic Labeling of Video Events with Verbs Based on
  Event-Participant Interaction}

\urldef\webpage\path{http://engineering.purdue.edu/~qobi/arxiv2012d}

\author{Andrei Barbu,$^{a}$
Alexander Bridge,$^{a}$
Dan Coroian,$^{a}$
Sven Dickinson,$^{b}$
Sam Mussman,$^{a}$\\
Siddharth Narayanaswamy,$^{a}$
Dhaval Salvi,$^{c}$
Lara Schmidt,$^{a}$
Jiangnan Shangguan,$^{a}$\\
Jeffrey Mark Siskind,$^{a}$\thanks{%
  \hspace{1ex} Corresponding author. Email: {\small\texttt{qobi@purdue.edu}}.}$\;$
Jarrell Waggoner,$^{c}$
Song Wang,$^{c}$
Jinlian Wei,$^{a}$
Yifan Yin,$^{a}$ and
Zhiqi Zhang$^{c}$}

\date{\begin{tabular}[t]{l}
$^{a}$School of Electrical \& Computer Engineering, Purdue University,
West Lafayette, IN, USA\\
$^{b}$Department of Computer Science, University of Toronto, Toronto,
CA\\
$^{c}$Department of Computer Science \& Engineering, University of
South Carolina, Columbia, SC, USA
\end{tabular}}

\maketitle

\let\thefootnote\relax\footnotetext{Additional images and videos as well as all code and datasets are available at \webpage.}

\begin{abstract}
We present an approach to labeling short video clips with English verbs as
event descriptions.
A key distinguishing aspect of this work is that it labels videos with verbs
that describe the spatiotemporal interaction between event participants, humans
and objects interacting with each other, abstracting away all object-class
information and fine-grained image characteristics, and relying solely on the
coarse-grained motion of the event participants.
We apply our approach to a large set of~22 distinct verb classes and a corpus
of~2,584 videos, yielding two surprising outcomes.
First, a classification accuracy of greater than~70\% on a 1-out-of-22 labeling
task and greater than~85\% on a variety of 1-out-of-10 subsets of this labeling
task is independent of the choice of which of two different time-series
classifiers we employ.
Second, we achieve this level of accuracy using a highly impoverished
intermediate representation consisting solely of the bounding boxes of one or
two event participants as a function of time.
This indicates that successful event recognition depends more on the choice of
appropriate features that characterize the linguistic invariants of the event
classes than on the particular classifier algorithms.
\end{abstract}

\section{Introduction}
\label{sec:introduction}

People describe observed visual events using verbs.
A common assumption in Linguistics \citep{Jackendoff83, Pinker89} is that verbs
typically characterize the interaction between event participants in terms of
the gross changing motion of these participants.
Object class and image characteristics of the participants are believed to be
largely irrelevant to determining the appropriate verb label for an event.
Participants simply fill \defoccur{roles} (such as \defoccur{agent} and
\defoccur{patient}) in the spatiotemporal structure of the event class described
by a verb.
For example, an event where one participant (the agent) \emph{picks up} another
participant (the patient) consists of a sequence of two subevents, where during
the first subevent the agent moves towards the patient while the patient is at
rest and during the second subevent the agent moves together with the patient
away from the original location of the patient.
It does not matter whether the agent is a human or a cat, or whether the patient
is a ball or a cup.
Moreover, the shapes, sizes, colors, textures, \etc{} of the participants are
irrelevant.
Additionally, only the gross motion characteristics are relevant; it is
irrelevant whether the participants grow, shrink, bend, vibrate, \etc{}
during a \emph{pick up} event.
The precise linear or angular velocities and accelerations are likewise
irrelevant.

The objective of this paper is to evaluate this Linguistic assumption and its
relevance to the computer-vision task of labeling video events with verbs.
In order to evaluate this hypothesis, we focus our attention on methods that
classify events solely on the basis of the gross changing motion of the event
participants.
In doing do, we often expressly discard other sources of information such as
object class, changing human body posture, and low-level image characteristics
such as shape, size, color, and texture.
We do this not because we believe that such information could not help event
recognition but rather to allow us to strongly evaluate the above hypothesis.
The surprising result of this endeavor is that gross changing motion of event
participants attains greater than 70\% accuracy on a 1-out-of-22 forced-choice
labeling task, significantly outperforming chance (4.5\%), and greater than
85\% accuracy on a variety of 1-out-of-10 subsets of this labeling task, again
significantly outperforming chance (10\%).

As this paper focusses on labeling video events with verbs, both the methods
and datasets commonly used in prior event-classification efforts are not
appropriate.
Such work typically classifies events using object and image characteristics
and fine-grained shape and motion features, such as spatiotemporal volumes
\citep{Blank2005,Rodriguez2008,Laptev2008} and tracked feature points
\citep{Liu2009,Schuldt2004,Wang2009}.
Moreover, many of the datasets commonly used in such work do not involve people
interacting with objects or other people and contain event classes that do not
depict common verbs.
For example, the distinctions between \texttt{wave1} and \texttt{wave2} or
\texttt{jump} and \texttt{pjump} in the \textsc{Weizmann} dataset
\citep{Blank2005} or the distinctions between \texttt{Golf-Swing-Back},
\texttt{Golf-Swing-Front}, and \texttt{Golf-Swing-Side}; \texttt{Kicking-Front}
and \texttt{Kicking-Side}; or \texttt{Swing-Bench} and \texttt{Swing-SideAngle}
in the \textsc{Sports Actions} dataset \citep{Rodriguez2008} do not correspond
to distinctions in verb semantics.
The event classes \texttt{side} and \texttt{jack} in the \textsc{Weizmann}
dataset, the event classes \texttt{Swing-Bench} and \texttt{Swing-SideAngle} in
the \textsc{Sports Actions} dataset, and the vast majority of the event classes
in the \textsc{ucf50} dataset \citep{Liu2009} (\eg\ \texttt{Basketball},
\texttt{Billiards}, \texttt{BreastStroke}, \texttt{CleanAndJerk},
\texttt{HorseRace}, \texttt{HulaHoop}, \texttt{MilitaryParade},
\texttt{TaiChi}, or \texttt{YoYo}, just to name a few) do not correspond to
verbs in any language.
The videos in the \textsc{kth} dataset \citep{Schuldt2004} do not reflect the
true meanings of any verbs, let alone \texttt{boxing} or \texttt{clapping} or
\texttt{waving} ones hands.
Typical actions in specialized domains like ballet (\cf\ the \textsc{Ballet}
dataset \citep{Wang2009}) are described by nouns, not verbs, and often are not
part of common lay vocabulary.
The distinction between the event classes \texttt{golf\_swing},
\texttt{tennis\_swing}, and \texttt{swing} in the \textsc{Youtube} dataset
\citep{Liu2009} reflect distinctions in event participants, not the semantics
of the verb \emph{swing}.

\cite{Siskind1996} presented a technique for labeling
video events with verbs based on the changing motion patterns of the event
participants.
However, they only applied their technique to a small number of event
classes (six) and a small corpus of thirty-six videos, six per class.
Moreover, they derived the changing motion patterns using a rudimentary
tracker that was specific to color and motion using background subtraction.
Thus the event participants were limited to people's hands interacting with
colored blocks in uncluttered desktop environments with static backgrounds.
In this paper, we employ the same technique for labeling video events with
verbs but extend it to a much larger number of event classes (twenty two) and
evaluate it on a much larger corpus of 2,584 videos ranging from~6 to~584 per
class.
Since the corpus used in the present effort exhibits a wide variety of natural
event participants in a wide variety of cluttered environments with
nonstationary backgrounds, this paper employs novel and more general-purpose
techniques for deriving the changing motion patterns.
Moreover, Siskind \& Morris used only one algorithmic method, namely hidden
Markov models (HMMs), to classify the time series that characterize the changing
motion patterns.
Thus one might conclude that the performance of this approach is somehow
dependent on this choice of classifier.
In this paper, we employ two distinct time-series classification methods,
namely HMMs and dynamic time warping (DTW) and demonstrate that both achieve
essentially identical performance.
Thus it appears that the strength of the approach results from the general
principle of classifying events based on gross changing motion patterns, not on
the algorithmic particulars.
Moreover, we demonstrate a surprising result.
Our front-end tracker abstracts each video as one or two moving axis-aligned
rectangles.
Despite such an extremely impoverished representation that passes only 4 or 8
small integers per frame between the front-end tracker and the back-end
time-series classifier, and the fact that all training and classification is
performed solely on this impoverished representation, both of our classifiers
attain greater than 70\% accuracy on a 1-out-of-22 forced-choice labeling task
and greater than 85\% accuracy on a variety 1-out-of-10 subsets of this task.
This supports the common assumption in Linguistics that the meanings of many
common verbs are sensitive only to gross changing motion patterns of the event
participants and not the object class or image characteristics of those
participants.

The paper is organized as follows.
Section~\ref{sec:corpus} describes the new corpus that we use for this effort.
Section~\ref{sec:tracking} describes the tracking methods that we employ to
abstract each video in this corpus to one or two moving axis-aligned rectangles.
Section~\ref{sec:classification} describes the feature vectors that we extract
from this impoverished representation and the particulars of the training and
classification paradigms.
Section~\ref{sec:results} describes our experimental results.
Section~\ref{sec:conclusion} concludes with a discussion of potential
improvements.

\section{The Mind's Eye Corpus}
\label{sec:corpus}

\nocite{DARPA2010a, DARPA2010b} 
As part of the Mind's Eye program, DARPA has
produced a video corpus that is specifically designed to support labeling of
videos with common verbs.
The particulars of this corpus were driven by the desire to ground the
semantics of 48 specific English verbs.
To date, several components of this corpus have been released to program
participants.
One portion, C-D1a, containing 2,584 videos, was released in late September
2010, while a second portion, C-D1b, containing 1,564 videos, was released in
late January 2011.
The videos are provided at 720p@30fps and range from 21 frames to 1408 frames in
length, with an average of 241 frames.
The videos in C-D1a range from 21 frames to 809 frames in length, with an
average of 141 frames.
Each video is intended to depict one of the 48 specific English verbs and
collectively all 48 verbs are represented in this combined corpus (with unequal
numbers of exemplar videos).
Each video comes labeled with the intended verb depiction.\footnote{Our corpus-
  size measurements reflect only the videos in the \texttt{SINGLE\_VERB}
  directory of C-D1a, and eliminate from consideration those videos not labeled
  with a single verb from this list of 48 verbs.}
Because verbs often exhibit a range of polysemous and homonymous meanings and
also may exhibit synonymy where the semantic space of one verb may include all
or part of the semantics space of another verb, DARPA intends to eventually
solicit human judgements for the association of verb labels with each video.
Since such human labelings have not yet been produced, in this paper we simply
take the `correct' label for each video to be the intended verb label provided
with the video.
Moreover, this paper considers only the C-D1a portion that depicts 22 specific
English verbs.
Fig.~\ref{fig:actions} summarizes the distribution of verbs and exemplar
videos in this portion of the corpus.

\begin{figure}
\begin{center}
  \begin{tabular}{lr|lr|lr}
    approach & 584 & drop     &  44 & leave  & 116 \\
    arrive   &   8 & exchange &  18 & lift   & 78  \\
    attach   &  48 & fall     & 134 & pass   & 76  \\
    bounce   &  22 & give     & 552 & pick-up & 40  \\
    catch    & 201 & go       &   6 & pull   & 8   \\
    chase    & 108 & jump     & 150 & run    & 76  \\
    collide  & 101 & kick     &  48 & throw  & 26  \\
    dig      & 140 &          &     &        &     \\
  \end{tabular}
\end{center}
\caption{The number of exemplar videos for each verb in the DARPA Mind's Eye
  C-D1a corpus.
  There are 2,584 videos and 22 verbs in total.}
\label{fig:actions}
\end{figure}

Conformant to the linguistic observation that object identity and class is
tangential to the task of labeling a video with a verb, different exemplars for
each of the verbs in C-D1a often have the participant roles played by different
object instances and classes.
The C-D1a corpus has a total of 26 distinct objects that play a role in the
depicted verbs as enumerated in Fig.~\ref{fig:objects}.
(Note that there are far more distinct objects that do not play a role in the
depicted verbs and serve solely to clutter the background.)
Many of these objects, however, only appear in the corpus occupying a very
small portion of the field of view and are difficult for humans, let alone
machines, to detect and classify reliably.
The ones that are most difficult to detect and classify reliably are
starred in Fig.~\ref{fig:objects}.
For each of the remaining ones, we manually cropped a collection of between
1,500 and 2,100 exemplars (combining both positive and negative samples) to
train a part-based object detector \citep{voc-release4,Felzenszwalb2010}.
It is important to stress that we use this object detector solely to produce
bounding-box information for deriving the gross changing motion patterns of the
event participants.
During event classification, we expressly discard the object-class information
and confidence scores provided by the object detector.
In section~\ref{sec:conclusion}, we discuss how one could extend our methods to
make use of such information and achieve even higher classification accuracy.

\begin{figure}
\begin{center}
  \begin{tabular}{@{}l@{\hspace*{7pt}}l@{\hspace*{7pt}}l@{\hspace*{7pt}}l@{}}
 bag&
 football$^{*}$ &
 rake \\
 bicycle &
 gun$^{*}$  &
 shovel \\
 big ball&
 hammer$^{*}$ &
 small ball\\
 bottle$^{*}$&
 keys$^{*}$ &
 spade$^{*}$\\
 bucket&
 log$^{*}$ &
 SUV \\
 cap$^{*}$&
 motorcycle&
 tape$^{*}$ \\
 cardboard box$^{*}$&
 pen$^{*}$ &
 tripod \\
 cellphone$^{*}$  &
 person   &
 wooden box\\
 chair  &
 pouch$^{*}$ &
  \end{tabular}
\end{center}
\caption{The 26 distinct objects that play a role in the depicted verbs in the
  C-D1a corpus.
The starred objects are the ones that are most difficult to detect and
classify reliably.}
\label{fig:objects}
\end{figure}

\section{Tracking}
\label{sec:tracking}

We use Felzenszwalb \etal's \citep{voc-release4, Felzenszwalb2010} part-based
object detector as a \defoccur{detection source} to produce axis-aligned
rectangles (henceforth \defoccur{detection boxes} or simply
\defoccur{detections} or \defoccur{boxes}) as a function of time.
However, it is unreliable alone as a means for characterizing gross
participant-object motion because it simultaneously exhibits a high
false-positive rate and a high false-negative rate.
Moreover, there is no single detection threshold that properly trades off the
false-positive and false-negative rates in a class- or video-independent
fashion.
Additionally, the raw detection-confidence values produced by the detector, or
even their rank ordering, cannot be used on isolated frames to select the
desired detection.
Moreover, the detector alone cannot distinguish between false positives and
multiple objects of the same class at close positions in the field of view.
Likewise, the detector alone does not provide temporal-correspondence
information in this situation.
These problems are particularly exacerbated by occlusion, where objects enter
and leave the field of view or pass in front of or behind other objects.
In these circumstances, the detection confidence becomes an even less reliable
measure of the (partially occluded) presence or absence of an object.
This is a particularly egregious limitation because verbs describe interaction
among participants and such interaction most frequently involves occlusion.

\subsection{Optimal selection of object tracks}
\label{sec:optimal}

We address all of these issues with a novel technique that produces coherent
\defoccur{object tracks} across a video from collections of independent
detections in each frame by simultaneously selecting among multiple detections
in all frames of a video to find the combination of selections that leads to a
global optimum of a cost function that characterizes the overall object-track
\defoccur{coherence}.
While we employ this technique using Felzenszwalb \etal's part-based object
detector as a detection source, it can be more generally applied to any
alternate detection source that outputs boxes with confidence scores.
The only requirement is that the confidence scores must provide a total
ordering of the boxes.
The confidence scores need not be normalized or lie in a particular interval.
This lax requirement facilitates integrating boxes produced by different
detection sources into a single coherent track, simply by providing a
correspondence between the confidence values produced by the different
detection sources and how they impact this total order.
We avail ourselves of this potential in section~\ref{sec:multiple} to provide
resilience in the face of appearance change due to nonrigid motion and
out-of-plane object rotation.

One can conceivably use an alternate detection source that does not rely on an
object detector.
For example, one might do some form of background subtraction or motion-based
tracking to separate moving objects from the background or some form of
bottom-up foreground/background segmentation or contour completion to segment
salient objects.
Any method that could reliably place bounding boxes around event participants
as a function of time would suffice for our purposes.
The sole reason that we employ an object detector as a detection source is that
bottom-up methods are currently not sufficiently reliable, while methods based
on background subtraction or motion detection fail to detect non-moving event
participants (of which there are many in our corpus) and are unreliable in the
presence of nonstationary backgrounds (such as occur frequently in our corpus).

We apply our detection source independently for each frame and each model,
biasing this detection source to yield few false negatives at the expense of
yielding a preponderance of false positives, and use our tracker to filter out
the false positives.
When using Felzenszwalb \etal's part-based object detector as a detection
source, we do this by subtracting a fixed offset (which we take to be~1) from
the learned detection threshold.
The particular value of this offset is unimportant so long as it yields a
sufficiently low false-negative rate, as our method reliably selects coherent
tracks despite an extremely high false-positive rate.
The only negative impact of choosing too high of an offset is an increase in
run time.

Felzenszwalb \etal's part-based object detector, by default, incorporates
non-maxima suppression to remove detections that overlap more than 50\% with
detections of higher confidence.
This tends to foil the above process for biasing the detector towards few false
negatives and many false positives.
To counter the effect of excessive non-maxima suppression, we raise the overlap
threshold to 80\%.
This allows for much better object localization and reduces jitter
considerably.

We have found that no amount of the above bias process will completely
eliminate false negatives.
To provide for robust production of coherent object tracks that are necessary
for successful event classification, we compensate for the remaining false
negatives by projecting each detection box in each frame forward a fixed number
of frames using the Kanade-Lucas-Tomasi (KLT) \citep{shi1994, tomasi1991}
feature tracker.
We track the KLT features that reside inside each detection box for one frame
and compute a single velocity vector and divergence vector for that detection
by computing the average velocity and divergence of the KLT features tracked
for that box.
We use the aggregate velocity and divergence vectors to project the detection
box forward one frame and repeat this process.
We limit this projection process to~5 frames as it is subject to drift, and we
need it only to compensate for false negatives which are relatively rare as a
result of the above bias process.
We augment the collection of detections to include the forward-projected boxes,
taking the confidence score of a forward-projected box to be that of the
original detection that was forward projected.

To select a coherent object track across multiple frames we construct a graph
with one vertex for each detection in each frame and edges connecting all pairs
of detections in adjacent frames.
The edges are weighted with a cost that inversely measures coherence and we
search for a path from the first to last frames with minimal total edge weight
using a dynamic-programming algorithm \citep{Viterbi1971} that finds a global
optimum.
This cost is formulated as a linear combination of two components, one being
the detection confidence score and the other being consistency with optical
flow.
The latter is taken to be the Euclidean distance between the center of a
detection box in a given frame and a projection of the center of the
corresponding detection box from the previous frame forward using optical flow.
This forward-projection process is analogous to the one performed to compensate
for false negatives except that the average velocity vector is computed from
dense optical flow instead of tracked KLT features.

In principle, one could use either KLT features or optical flow for either
forward-projection process.
We find that, in practice, KLT features yield better results for the
forward-projection process used to compensate for false negatives while optical
flow yields better results for the forward-projection process used to compute
track coherence.
Also, our track-coherence measure uses only the distance between detection-box
centers and thus does not need a divergence measure.
While one could extend the track-coherence measure to incorporate such
information, we find that it yields no improvement in performance.
In our experiments, we weight the optical-flow component of track coherence ten
times less than the detection-confidence score.
We bias the track-coherence measure towards detection confidence to prevent
production of tracks that are consistent with optical flow but do not
correspond to reliable object detections.
Other than this general bias, we find that the object tracks produced are
largely insensitive to the precise weighting value.

\subsection{Entering and leaving the field of view}
\label{sec:enter}

The algorithm described thus far constructs tracks that span the entire video
from the first frame to the last frame.
We allow for objects that enter and leave the field of view simply by applying
this algorithm to a subinterval of the video.
The only difficulty in doing so is determining the subinterval boundaries.
We take the subinterval to begin at the first frame with a detection confidence
above a certain threshold, and end at the last such frame.
To derive this threshold, we compute a (50 bin) histogram of the maximal
detection-confidence scores in each frame, over the entire video.
One expects this histogram to be bimodal since frames in which the object is
not present will have lower confidence scores, as all detections will be false
positives.
We take the threshold to be the minimum of the value that maximizes the
between-class variance \citep{Otsu1979} when bipartitioning this histogram and
the learned detector-confidence threshold offset by a fixed, but small, amount
(0.4).
In practice, we find that proper selection of the subinterval is largely
insensitive to the number of bins and the precise threshold offset.

\subsection{Multiple instances of the same object class}

We detect multiple tracks of the same object class by repeated application of
the above method.
In doing so, we must prevent subsequent iterations from rediscovering tracks
produced by earlier iterations.
The na\"{i}ve way of doing this would be to remove detections associated with
earlier tracks.
Detection boxes can be deemed to be associated with earlier tracks when their
centers lie inside detection boxes included in those earlier tracks.
However, removing all such detections runs the risk of precluding
overlapping tracks, as would happen when objects pass each other in the field
of view.
So instead of removing detections, we rescore them with the maximal
detection score in the lower quartile of scores for that frame.
Given the biasing process towards false positives away from false negatives in
the detection source, boxes in the lower quartile are likely to be false
positives and undesirable to include in a coherent track.
Rescoring detections in this fashion biases subsequent iterations to find
distinct tracks while allowing tracks to briefly overlap.

If one is not careful, there can be crossover at such points of overlap, where
the object identity is swapped between two distinct tracks.
We use an object-appearance model to bias against such crossover.
Color histograms are computed in the CIELAB \citep{CIE1978} color space of the
pixel values inside the detection boxes after shrinking those boxes by 60\% to
ameliorate the influence of background pixels on these histograms.
We then augment the edge-weight function to include a coherence measure on
object appearance, taking this coherence measure to be Earth Mover's distance
\citep{Peleg1989} between the corresponding histograms.
We weight object appearance and detector confidence equally in the coherence
measure, though in practice, we find that the object tracks produced are
largely insensitive to the precise weighting.

\subsection{Nonrigid motion and out-of-plane rotation}
\label{sec:multiple}

Felzenszwalb \etal's part-based object detector is unreliable as a detection
source when there is nonrigid motion and out-of-plane rotation.
Our tracking framework can provide resilience in the face of such unreliability
by integrating detection boxes from multiple detection sources.
We do so by training multiple models for Felzenszwalb \etal's part-based object
detector for varying object appearance under nonrigid motion and out-of-plane
rotation and union the resulting detections.
As discussed in section~\ref{sec:optimal}, we must insure that the confidence
scores allow for comparison between detections produced by different detection
sources.
We do this by offsetting the confidence scores for each detection source by the
threshold computed in section~\ref{sec:enter}.

The C-D1a corpus has little out-of-plane rotation and therefore such does not
impact the reliability of the detection source.
But the corpus does contain one source of nonrigid motion, namely changing
human body posture.
For this corpus, it is sufficient to train detectors for three distinct
postures: standing, crouching, and lying down.

Integrating multiple detection sources into a single object track allows
annotation of the detections in that track with their source.
In particular, this allows temporal annotation of human motion tracks with
their changing posture.
Conceivably one could use such information to support selection of the
appropriate verb label.
Because we wish to evaluate the hypothesis that verbs typically characterize
the gross changing motion of the event participants, we expressly discard such
information in the experiments performed in this paper.

\subsection{Smoothing}

Boxes comprising the recovered object tracks suffer from jitter.
We remove this jitter by fitting piecewise cubic splines to the
widths, heights, and~$x$ and~$y$ center coordinates of the tracked boxes.
A simple selection of smoothing parameters suffices for the C-D1a corpus.
Since the videos in C-D1a have low frame length variance, a constant number of
spline pieces is adequate.
Box~$x$ and~$y$ center coordinates are smoothed with~10 pieces, as they can
move significantly when tracking accelerating objects, for example a bouncing
ball.
Box widths and heights are smoothed with~5 pieces as object shape and size
change less drastically.

\subsection{Results}

\begin{figure*}
\begin{tabular}{@{}c@{\hspace*{6.5pt}}c@{}}
\includegraphics[width=0.46\textwidth]{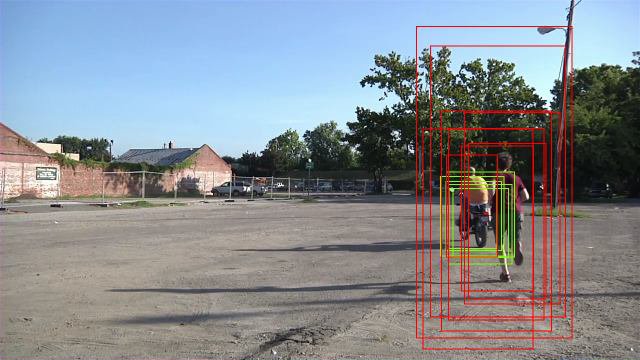}&
\includegraphics[width=0.46\textwidth]{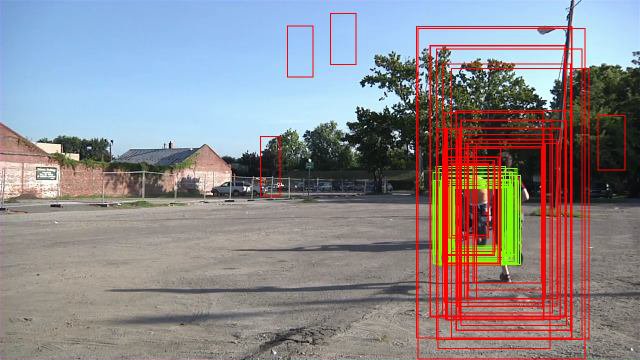}\\
(a)&(b)\\
\includegraphics[width=0.46\textwidth]{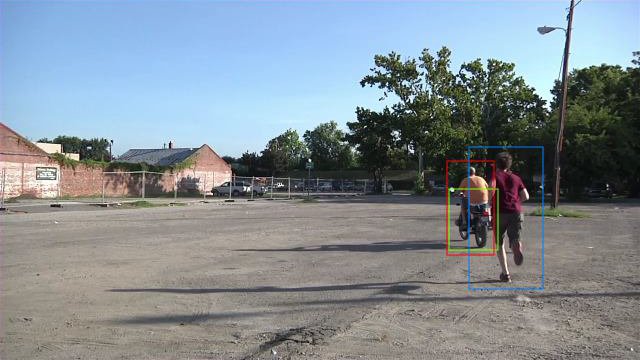}&
\includegraphics[width=0.46\textwidth]{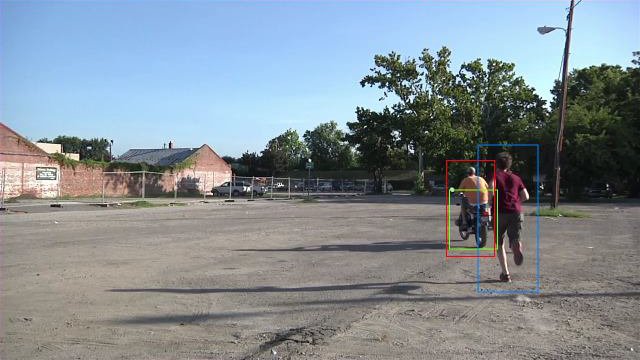}\\
(c)&(d)\\
\end{tabular}
\caption{The output of each stage of our tracker on a single frame.
  (a)~Detections for the person model in red and the motorcycle model in
  green.
  (b)~Forward projections of the detections from the~5 previous frames.
  (c)~The object tracks with maximal coherence selected by our
  dynamic-programming algorithm.
  Two distinct person tracks are shown in red and blue, and a motorcycle track
  is shown in green.
  (d)~The smoothed tracks.}
\label{fig:tiled-tracker}
\end{figure*}

\begin{figure*}
\begin{tabular}{@{}c@{\hspace*{6.5pt}}c@{}}
\includegraphics[width=0.46\textwidth]{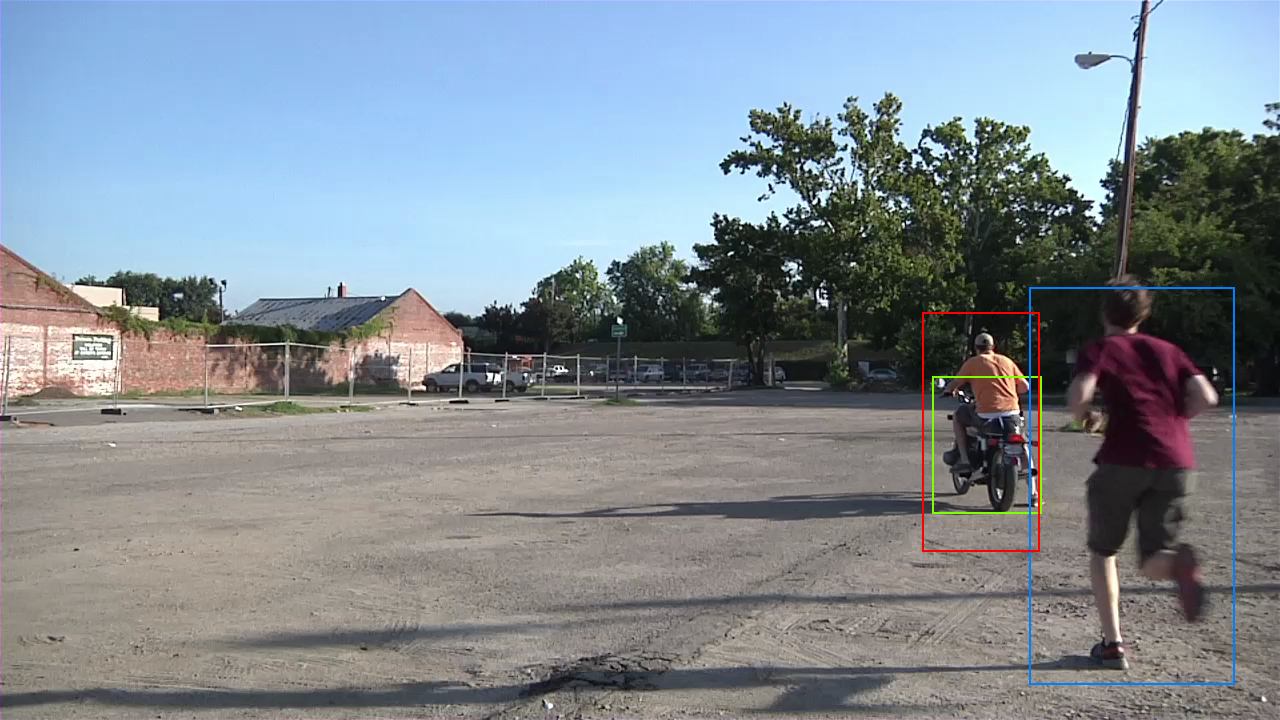}&
\includegraphics[width=0.46\textwidth]{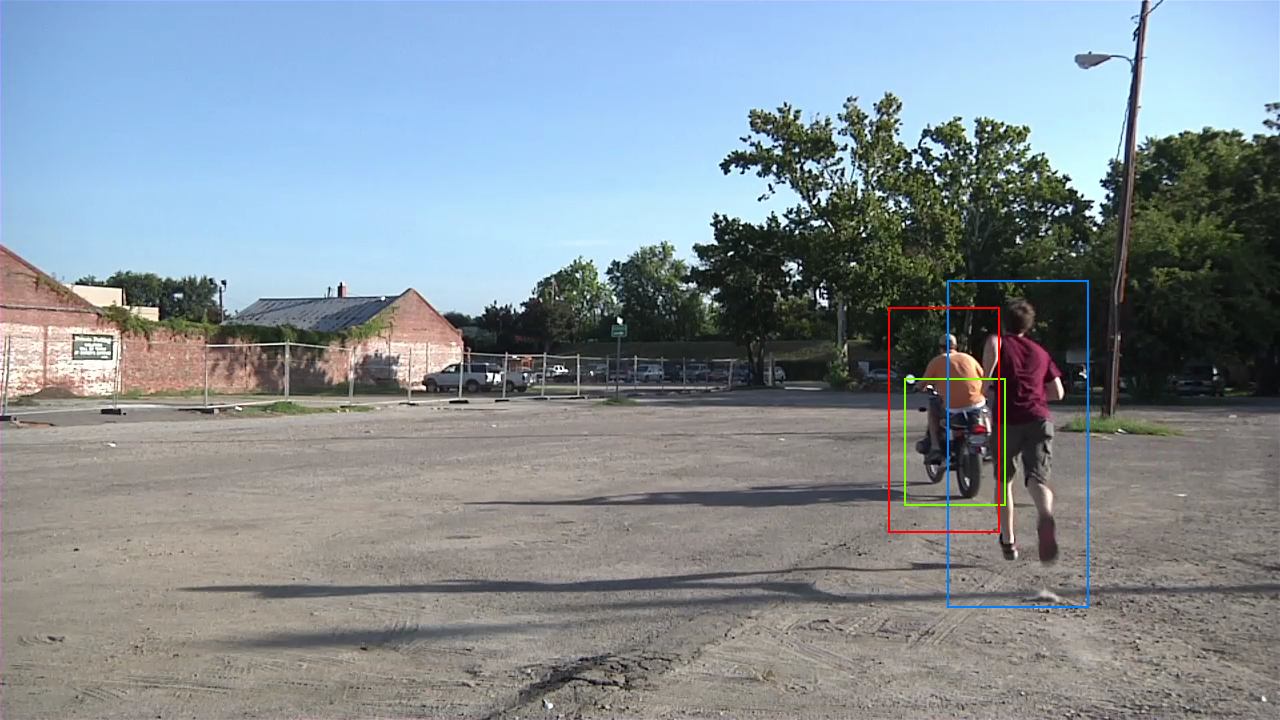}\\
\includegraphics[width=0.46\textwidth]{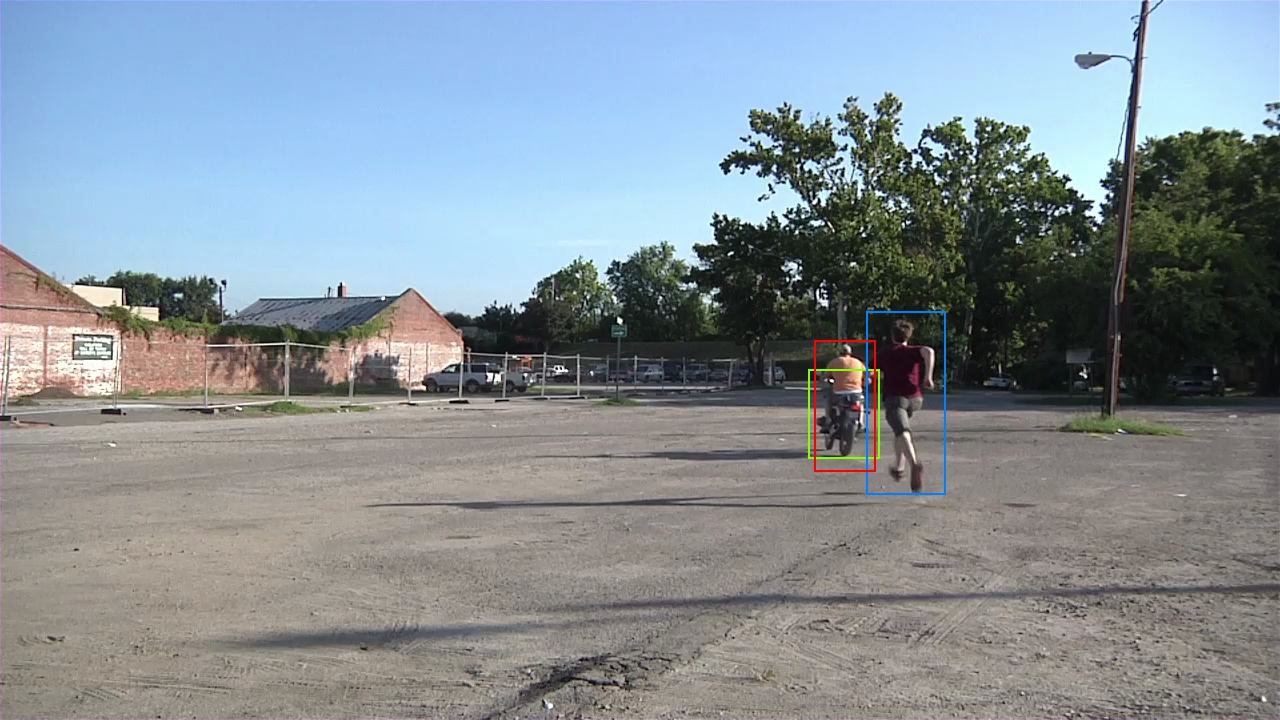}&
\includegraphics[width=0.46\textwidth]{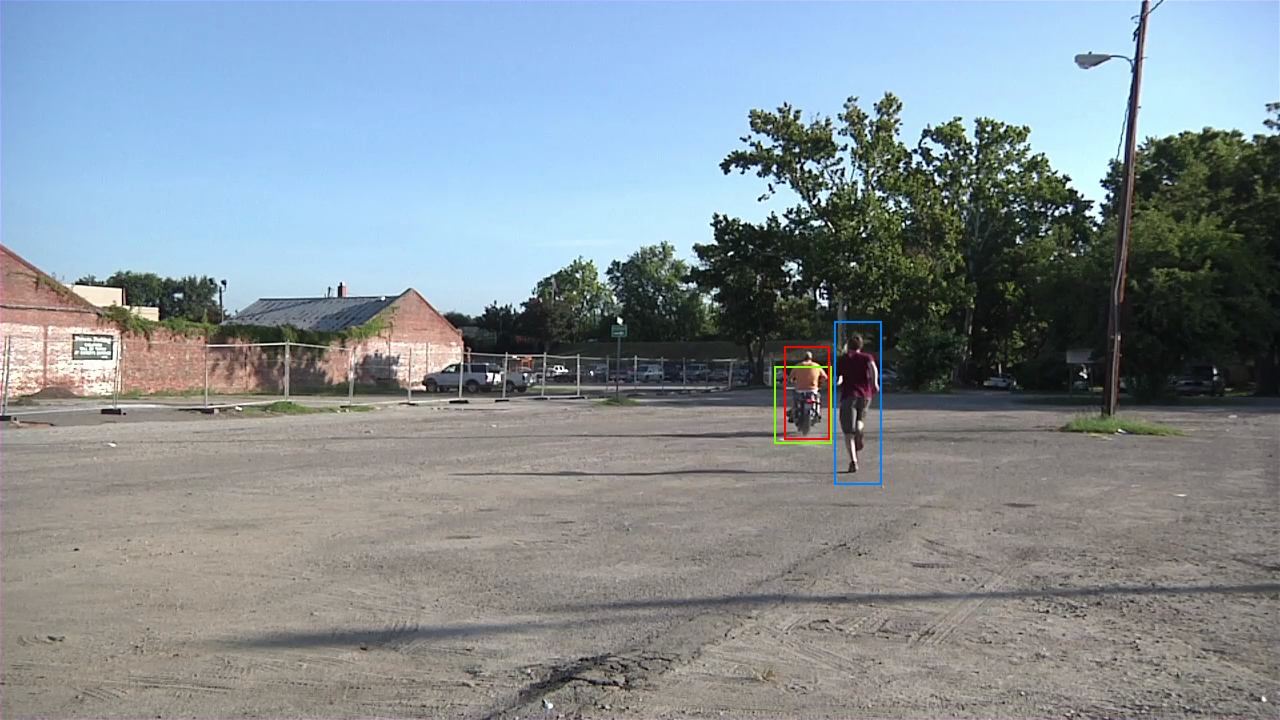}\\
\end{tabular}
\caption{Four frames tracking two people and one motorcycle.
Two separate person tracks in red and blue, and the motorcycle track in green.
The tracker is robust despite the fact that one person occludes most of the
motorcycle, the tracks of the two people overlap, and all three objects become
very small as they recede from the camera.}
\label{fig:tracker-images}
\end{figure*}

Our tracker runs in time $O(lm+lmn|df|^2)$ to recover~$n$ tracks with~$m$
detection sources, each yielding~$d$ detections per frame, doing~$f$ frames of
forward projection, on videos of length~$l$.
In practise, the run time is dominated by the detection process and the
dynamic-programming step.
Fig.~\ref{fig:tiled-tracker} illustrates the operation of our tracker,
rendering the output of each stage.
From this video, one can clearly see the robustness of our tracker in light
of cluttered nonstationary backgrounds, motion that is not perpendicular to the
camera axis, an extremely high false-positive biased detection rate of the
detection source, occlusion that results from overlapping tracks corresponding
to interacting objects, nonrigid motion that results from changing human body
posture, objects entering and leaving the field of view, and multiple instances
of the same object class.
Moreover, as illustrated in Fig.~\ref{fig:tracker-images}, the fact that our
tracker finds an optimal coherent track by processing the entire video allows it
to robustly track objects that approach or recede from the camera by a large
distance that would otherwise be too small in the field of view to reliably
track by methods that did not process the entire video.
Without the false-positive bias that such a whole-video approach allows,
Felzenszwalb \etal's part-based object detector would not even detect such
objects.

\section{Classification}
\label{sec:classification}

We convert the collection of object tracks for a video to a time-series of
real-valued feature vectors and formulate the problem of labeling a video with
a verb as a time-series classification problem.
In doing so, we discard all object identity and body posture information that is
available in those tracks.

For each video, we designate one track as the agent and another track (if
present) as the patient.
The agent is determined using a heuristic: people are more likely to be agents
than inanimate objects are, and bicycles, motorcycles, and SUVs are more
likely to be agents than other inanimate objects because they are driven by
people that we might fail to detect due to occlusion.
Another track (if present) is selected as the patient using the same heuristic.
Ties are broken by selecting the track with highest track coherence
as the agent and the one with second highest track coherence as the patient.

For all videos, we extract a feature vector for each frame representing the
gross absolute motion of the agent:
\begin{compactenum}
\item $x$-coordinate of the box center
\item $y$-coordinate of the box center
\item box aspect ratio
\item derivative of the box  aspect ratio
\item magnitude of the velocity of the box center
\item direction of the velocity of the box center
\item magnitude of the acceleration of the box center
\item direction of the acceleration of the box center
\end{compactenum}
For videos with two or more object tracks, we also extract a feature vector
that includes the above absolute motion features representing the independent
motion of each of the agent and patient along with additional features that
describe their relative motion:
\begin{compactenum}
\item distance between agent and patient box centers
\item orientation of vector from agent box center to patient box center
\item derivative of the distance between agent and patient box centers
\end{compactenum}
In all of the above, temporal derivatives and corresponding velocities and
accelerations are computed as a two-point finite difference.
Note that we label videos with verbs using the gross changing motion patterns
of at most two event participants.
While we could, in principle, label videos on the basis of the motion patterns
of more event participants, if present, by straightforward extension of the
above feature-vector computation to include absolute features for all objects
and relative features for all object pairs, we expressly refrain from doing so
to evaluate the Linguistic hypothesis that verbs largely describe the
interaction between an agent and a patient.

The verbs in C-D1a often have different senses, such as the causative/inchoative
alternation (the agent \emph{bounces} \vs{} the agent \emph{bounces} the
patient), that involve a different number of participants.
In this case, we train two distinct classifiers, one on all videos
characterizing the motion of just the agent and one on those videos that have
both an agent and a patient characterizing the motion of both the agent and the
patient.
When classifying an unseen video with just a single object track we use models
trained on just agents, while when classifying an unseen video with more than
one object track we use models trained on both agents and patients.

To evaluate the hypothesis that it is possible to classify events solely on
the basis of the gross changing motion of the event participants and
demonstrate the insensitivity of this hypothesis to the choice of time-series
classifier, we have run two parallel sets of experiments, one with HMMs
\citep{Baum1966} and one with DTW \citep{Ellis2003, Sakoe1978}.
When using HMMs, we train models with~5 states and independent continuous output
distributions for each feature.
We use Gaussian distributions for those features that constitute linear
quantities and Von Mises distributions for those features that constitute
angular quantities.
We found that increasing the number of states beyond~5 did not significantly
improve accuracy.
When using DTW, we employ Euclidean distance between feature vectors as the
distance metric between frames and use DTW to extend this metric as a
distance between frame sequences to construct a nearest-neighbor classifier
between unseen videos and training exemplars.

\section{Results}
\label{sec:results}

We performed 5-fold cross-validation on the entire C-D1a corpus with a
1-out-of-22 forced-choice classification task using both HMMs and DTW.\@
To do this, we independently partitioned the set of `correct' exemplars for each
verb into five random but equally sized components (up to quantization).
For each of the five cross-validation runs we trained on the exemplars in four
of the five partitions and tested on the exemplars in the remaining partition.
Fig.~\ref{fig:5-fold} gives the recognition accuracy for each classification
algorithm for each cross-validation run.
Fig.~\ref{fig:confusion-matrix-hmm} and Fig.~\ref{fig:confusion-matrix-dtw}
give the aggregate confusion matrices for each classification algorithm across
all five cross-validation runs.
Note the essentially identical performance of HMMs and DTW: HMMs exhibits an
aggregate classification accuracy of 71.9\% while DTW exhibits an aggregate
classification accuracy of 71.3\%.
Moreover, we attain greater than 85\% aggregate classification accuracy for
three different 1-out-of-10 subsets of this forced-choice classification task
with both HMMs and DTW:
\emph{arrive bounce dig drop exchange give jump kick pickup run}
(87.4\% HMMs, 85.3\% DTW),
\emph{bounce dig drop exchange give jump kick pickup pull run}
(87.5\% HMMs, 85.1\% DTW), and
\emph{bounce dig drop exchange give jump kick pass pickup pull}
(86.1\% HMMs, 87.0\% DTW).
These results support the hypothesis that classification accuracy depends more
on the correct choice of features than on the classification algorithm.

\begin{figure}
\vspace{3ex}
  \begin{center}
  \begin{tabular}{ c r r r r r }
    HMMs & 74.4 & 72.9 & 70.5 & 69.5 & 72.4  \\
    DTW & 70.7 & 71.2 & 69.5 & 75.0 & 70.2 \\
  \end{tabular}
  \end{center}
\caption{Accuracy for HMMs and DTW on the 1-out-of-22 action classification
  task for each of the 5 random partitions of the corpus.}
\label{fig:5-fold}
\end{figure}

\section{Conclusion}
\label{sec:conclusion}

Our focus in this paper is to evaluate the hypothesis that it is possible to
label videos with verbs using information solely about the gross changing
motion of the event participants.
There are numerous places where our computational methods expressly discard
information that is otherwise available in order to evaluate this hypothesis.
Since such information might correlate with the underlying event, one could
extend our classifiers to make use of such information.
For example, one might expect that detector confidence scores would decrease
with occlusion and thus correlate with the object interaction indicative of
event class.
Similarly, one might expect that object class would correlate with event
class.
Indeed, as shown in Fig.~\ref{fig:objectclass}, such correlation significantly
reduces the potential verb-label space, rendering the verb-labeling task almost
trivial.
Likewise, as discussed in section~\ref{sec:multiple}, one could augment the time
series of feature vectors with human body-posture information that is extracted
as a by-product of using multiple detection sources to provide resilience in
the face of out-of-plane rotation and nonrigid motion.
It is quite unexpected that we attain as good results as we have despite
expressly discarding such information.
This supports the common assumption in Linguistics that verbs typically
characterize the interaction between event participants in terms of the gross
changing motion of these participants.

\begin{figure}
\begin{eqnarray*}
\textrm{bag}              &\rightarrow& \textit{lift}\\
\textrm{bicycle}          &\rightarrow& \textit{give}\\
\textrm{big ball}    &\rightarrow& \textit{appoach} \;|\; \textit{chase} \;|\; \textit{catch} \;|\; \textit{collide}\\
\textrm{bucket}           &\rightarrow& \textit{dig}\\
\textrm{chair}            &\rightarrow& \textit{give} \;|\; \textit{collide} \;|\; \textit{fall}\\
\textrm{football}         &\rightarrow& \textit{catch} \;|\; \textit{throw}\\
\textrm{motorbike}        &\rightarrow& \textit{give} \;|\; \textit{approach} \;|\; \textit{chase} \;|\; \textit{leave} \;|\; \textit{run}\\
\textrm{rake}             &\rightarrow& \textit{dig}\\
\textrm{shovel}           &\rightarrow& \textit{dig}\\
\textrm{small ball} &\rightarrow& \textit{collide} \;|\; \textit{lift}\\
\textrm{SUV}              &\rightarrow& \textit{give} \;|\; \textit{approach} \;|\; \textit{chase} \;|\; \textit{leave} \;|\; \textit{catch}\\
                          &\;|\;& \textit{throw} \;|\; \textit{run}\\
\textrm{wooden box}       &\rightarrow& \textit{give}
\end{eqnarray*}
\caption{Correlation between object and event class in C-D1a.}
\label{fig:objectclass}
\end{figure}

\section*{Acknowledgments}

This work was supported, in part, by NSF grant CCF-0438806, by the Naval
Research Laboratory under Contract Number N00173-10-1-G023, by the Army
Research Laboratory accomplished under Cooperative Agreement Number
W911NF-10-2-0060, and by computational resources provided by Information
Technology at Purdue through its Rosen Center for Advanced Computing.
Any views, opinions, findings, conclusions, or recommendations contained or
expressed in this document or material are those of the author(s) and do not
necessarily reflect or represent the views or official policies, either
expressed or implied, of NSF, the Naval Research Laboratory, the Office of
Naval Research, the Army Research Laboratory, or the U.S. Government.
The U.S. Government is authorized to reproduce and distribute reprints for
Government purposes, notwithstanding any copyright notation herein.

\bibliographystyle{plainnat}
\bibliography{arxiv2012d}

\onecolumn

\begin{landscape}
\begin{figure*}
\begin{center}
\resizebox{1.3\textwidth}{!}{
\begin{tabular}{ r| r r r r r r r r r r r r r r r r r r r r r r }
 & \emph{approach} & \emph{arrive} & \emph{attach} & \emph{bounce} & \emph{catch} & \emph{chase} & \emph{collide} & \emph{dig} & \emph{drop} & \emph{exchange} & \emph{fall} & \emph{give} & \emph{go} & \emph{jump} & \emph{kick} & \emph{leave} & \emph{lift} & \emph{pass} & \emph{pick-up} & \emph{pull} & \emph{run} & \emph{throw}\\\hline
\emph{approach} & {\green  87} &   &   2 &   &   &   4 &   4 &   2 &   &   6 &   &   5 &  17 &   &   &   3 &   &   5 &   &   &   &  \\
\emph{arrive} &   & {\green  63} &   &   &   &   &   &   &   &   &   &   &   &   &   &   &   &   &   &   &   &  \\
\emph{attach} &   &   & {\green  69} &   &  12 &   &   &   1 &   &   &   3 &   2 &   &   1 &   &   &   1 &   &   &   &   &  \\
\emph{bounce} &   &   &   & {\green  91} &   2 &   2 &   &   &   &   &   1 &   &   &   &   &   &   &   &   &   &   &  \\
\emph{catch} &   &   &   8 &   & {\green  54} &   &   1 &   &   &  28 &   4 &   3 &   &   &   &   &   5 &   &   5 &   &   &  \\
\emph{chase} &   2 &   &   2 &   &   4 & {\green  81} &   5 &   &   2 &   &   5 &   1 &  17 &   1 &   &   6 &   1 &   5 &   &  13 &   5 &   8\\
\emph{collide} &   1 &   &   &   9 &   3 &   2 & {\green  66} &   1 &   &   &   3 &   2 &   &   1 &   &  11 &   1 &  14 &   &   &   1 &  \\
\emph{dig} &   1 &   &   4 &   &   &   &   & {\green  90} &   &   &   &   1 &   &   &   &   &   &   &   &   &   1 &  \\
\emph{drop} &   &   &   &   &   1 &   &   &   & {\green  61} &   &   &   2 &   &   1 &   &   1 &   9 &   &  10 &   &   1 &   4\\
\emph{exchange} &   1 &   &   4 &   &   5 &   &   2 &   &   & {\green  28} &   &   1 &   &   &   &   1 &   &   &   &   &   &  \\
\emph{fall} &   1 &   &   &   &   3 &   &   3 &   &   &   & {\green  77} &  12 &   &   1 &   &   2 &   &   &   &   &   3 &  \\
\emph{give} &   1 &   &   2 &   &   &   4 &   2 &   1 &   5 &  17 &   3 & {\green  56} &   &   &   &   3 &   6 &   &   &   &   &  \\
\emph{go} &   &   &   &   &   &   &   &   &   &   &   &   & {\green  17} &   &   &   1 &   &   &   &   &   &  \\
\emph{jump} &   &   &   &   &   2 &   1 &   1 &   &   &   &   1 &   1 &   & {\green  90} &   &   4 &   3 &   &   &   &   3 &  \\
\emph{kick} &   &   &   &   &   1 &   &   &   &   &   &   &   1 &   &   2 & {\green 100} &   &   1 &   &   &   &   &  12\\
\emph{leave} &   2 &  38 &   &   &   &   4 &   &   1 &   &  22 &   &   3 &  50 &   &   & {\green  45} &   &   1 &   &  13 &   1 &  \\
\emph{lift} &   1 &   &   &   &   &   1 &   &   4 &  27 &   &   &   3 &   &   1 &   &   3 & {\green  65} &   &   5 &   &   1 &  12\\
\emph{pass} &   2 &   &   &   &   &   1 &  16 &   1 &   &   &   1 &   3 &   &   3 &   &   9 &   & {\green  74} &   &   &   4 &  \\
\emph{pick-up} &   &   &   6 &   &   1 &   &   &   &   5 &   &   1 &   &   &   &   &   &   4 &   & {\green  80} &   &   &  \\
\emph{pull} &   &   &   &   &   &   &   &   &   &   &   &   &   &   &   &   &   &   &   & {\green  75} &   1 &  \\
\emph{run} &   &   &   &   &   &   1 &   &   &   &   &   1 &   1 &   &   &   &  11 &   &   &   &   & {\green  78} &   4\\
\emph{throw} &   1 &   &   2 &   &   7 &   &   &   &   &   &   1 &   3 &   &   &   &   &   3 &   &   &   &   & {\green  62}\\
\end{tabular}}
\end{center}
\caption{The aggregate confusion matrices for 5-fold cross validation on
  the 1-out-of-22 classification task using HMMs.
The overall accuracy is 71.9\%.}
\label{fig:confusion-matrix-hmm}
\end{figure*}
\end{landscape}

\begin{landscape}
\begin{figure*}
\begin{center}
\resizebox{1.3\textwidth}{!}{
\begin{tabular}{ r| r r r r r r r r r r r r r r r r r r r r r r }
 & \emph{approach} & \emph{arrive} & \emph{attach} & \emph{bounce} & \emph{catch} & \emph{chase} & \emph{collide} & \emph{dig} & \emph{drop} & \emph{exchange} & \emph{fall} & \emph{give} & \emph{go} & \emph{jump} & \emph{kick} & \emph{leave} & \emph{lift} & \emph{pass} & \emph{pick-up} & \emph{pull} & \emph{run} & \emph{throw}\\\hline
\emph{approach} & {\green  72} &   &   8 &   &   7 &   4 &  10 &  10 &   5 &  17 &   8 &   8 &   &   3 &   &   8 &  14 &  11 &   5 &  38 &   5 &   8\\
\emph{arrive} &   & {\green  50} &   &   &   1 &   &   &   &   &   &   &   &   &   &   &   &   1 &   &   &   &   &  \\
\emph{attach} &   1 &   & {\green  73} &   &   2 &   &   &   1 &   &   &   &   1 &   &   &   &   1 &   &   &   &   &   1 &  \\
\emph{bounce} &   &   &   & {\green  86} &   1 &   4 &   &   &   &   &   &   &   &   &   &   1 &   &   &   &   &   &  \\
\emph{catch} &   2 &  13 &   &   & {\green  71} &   &   &   4 &   &   &   3 &   1 &   &   &   &   &   3 &   &   &   &   &   8\\
\emph{chase} &   &   &   &   9 &   & {\green  71} &   4 &   1 &   5 &  17 &   &   &  67 &   1 &   &   8 &   &   &   &   &   4 &  \\
\emph{collide} &   2 &   &   &   &   1 &   2 & {\green  62} &   1 &   &   &   4 &   1 &   &   1 &   2 &   7 &   &   4 &   &   &   &  \\
\emph{dig} &   &   &   &   &   &   &   & {\green  52} &   2 &   &   1 &   &   &   &   &   1 &   &   &   &   &   3 &  \\
\emph{drop} &   &   &   &   &   &   &   1 &   2 & {\green  73} &   &   &   &   &   &   2 &   &   &   &   &   &   &   4\\
\emph{exchange} &   1 &   &   &   &   &   &   1 &   1 &   & {\green  28} &   &   &   &   &   &   &   &   &   &   &   &  \\
\emph{fall} &   1 &   &   &   &   4 &   2 &   2 &   1 &   &   6 & {\green  60} &   2 &   &   1 &   2 &   5 &   &   &   &   &   &  12\\
\emph{give} &  12 &   &  10 &   5 &  10 &   6 &   4 &  13 &   5 &  17 &  13 & {\green  82} &   &   1 &   4 &   4 &   8 &   3 &   5 &  25 &   7 &   4\\
\emph{go} &   &   &   &   &   &   3 &   &   &   &   &   &   & {\green   0} &   &   &   1 &   &   &   &   &   &   4\\
\emph{jump} &   1 &   &   &   &   &   2 &   2 &   1 &   &   &   2 &   &   & {\green  88} &   6 &   &   1 &   1 &   5 &   &   7 &  \\
\emph{kick} &   1 &   &   &   &   1 &   &   1 &   &   &   &   1 &   1 &   &   1 & {\green  83} &   &   &   1 &   &   &   &  \\
\emph{leave} &   1 &  13 &   2 &   &   1 &   5 &   2 &   1 &   5 &  11 &   1 &   &  17 &   2 &   & {\green  57} &   &   1 &   &  13 &   5 &   4\\
\emph{lift} &   2 &   &   4 &   &   2 &   &   &   1 &   2 &   &   1 &   2 &   &   1 &   &   1 & {\green  71} &   1 &  13 &  13 &   1 &  \\
\emph{pass} &   1 &   &   &   &   &   &   8 &   2 &   &   &   1 &   &   &   1 &   &   3 &   & {\green  76} &   &   &   &  \\
\emph{pick-up} &   2 &   &   &   &   &   &   &   3 &   2 &   &   &   &   &   &   &   &   1 &   & {\green  68} &   &   1 &  \\
\emph{pull} &   &   &   &   &   1 &   &   &   &   &   &   &   &   &   &   &   &   &   &   & {\green  13} &   &  \\
\emph{run} &   1 &   &   2 &   &   &   1 &   2 &   2 &   &   &   3 &   1 &  17 &   1 &   &   4 &   &   &   5 &   & {\green  66} &   4\\
\emph{throw} &   &  25 &   &   &   2 &   1 &   1 &   2 &   2 &   6 &   &   &   &   &   &   &   1 &   1 &   &   &   & {\green  54}\\
\end{tabular}
}
\end{center}
\caption{The aggregate confusion matrices for 5-fold cross validation on
  the 1-out-of-22 classification task using DTW.
The overall accuracy is 71.3\%.}
\label{fig:confusion-matrix-dtw}
\end{figure*}
\end{landscape}

\end{document}